\documentclass[16pt]{article}
\usepackage{comment}
\usepackage[paperheight=11.5in,paperwidth=9.5in]{geometry}

\usepackage{pdflscape}
\usepackage{graphicx} 
\usepackage[utf8]{inputenc} 
\usepackage[T1]{fontenc}    
\usepackage{hyperref}       
\usepackage{url}            
\usepackage{booktabs}       
\usepackage{amsfonts}       
\usepackage{nicefrac}       
\usepackage{microtype}      
\usepackage{lipsum}
\usepackage{color}
\usepackage{float}

\title{Embodied AI-Driven Operation of Smart Cities: A Concise Review}

\author{Farzan Shenavarmasouleh$^1$, Farid Ghareh Mohammadi$^1$, \\ M. Hadi Amini$^2$, and Hamid R. Arabnia$^1$\vspace{0.25in}\\ 1:Department of Computer Science, Franklin College of arts and sciences,\\ University of Georgia, Athens, GA, USA  \\
2: School of Computing \& Information Sciences, College of Engineering \& Computing, \\Florida International University, Miami, FL, USA \\
Emails: fs04199@uga.edu, farid.ghm@uga.edu, amini@cs.fiu.edu, hra@uga.edu}
      
\begin{document}
\maketitle

\begin{abstract}
A smart city can be seen as a framework, comprised of Information and Communication Technologies (ICT). An intelligent network of connected devices that collect data with their sensors and transmit them using wireless and cloud technologies in order to communicate with other assets in the ecosystem plays a pivotal role in this framework. Maximizing the quality of life of citizens, making better use of available resources, cutting costs, and improving sustainability are the ultimate goals that a smart city is after. Hence, data collected from these connected devices will continuously get thoroughly analyzed to gain better insights into the services that are being offered across the city; with this goal in mind that they can be used to make the whole system more efficient.
Robots and physical machines are inseparable parts of a smart city. Embodied AI is the field of study that takes a deeper look into these and explores how they can fit into real-world environments. It focuses on learning through interaction with the surrounding environment, as opposed to Internet AI which tries to learn from static datasets. Embodied AI aims to train an agent that can See (Computer Vision), Talk (NLP), Navigate and Interact with its environment (Reinforcement Learning), and Reason (General Intelligence), all at the same time. Autonomous driving cars and personal companions are some of the examples that benefit from Embodied AI nowadays.
In this paper, we attempt to do a concise review of this field. We will go through its definitions, its characteristics, and its current achievements along with different algorithms, approaches, and solutions that are being used in different components of it (e.g. Vision, NLP, RL). We will then explore all the available simulators and 3D interactable databases that will make the research in this area feasible. Finally, we will address its challenges and identify its potentials for future research.
\end{abstract}

\textbf{keywords:} Embodied AI, Embodied Intelligence, Question Answering, Smart Cities, Simulation, Intelligence, Multi-Agent Systems

\section{Introduction}
A smart city is an urban area that employs Information and Communication Technologies (ICT) \cite{park2021emerging} , an intelligent network of connected devices and sensors that can work interdependently \cite{hadi2_interdependent, mohammadi2019promises} and a distributive manner \cite{hadi3_distributed} to continuously monitor the environment, collect data, and share them among the other assets in the ecosystem. A smart city uses all the available data to make real-time decisions about the many individual components of the city to ease up the livelihood of its citizens, and make the whole system more efficient, more environmentally friendly, and more sustainable \cite{hadi5_sustainable}.
This serves as a catalyst for creating a city with faster transportation, fewer accidents, enhanced manufacturing, more reliable medical services and utilities, less pollution \cite{hadi1_pollution} and much more.
The good news is any city, even with traditional infrastructures, can be transformed into a Smart City by integrating IoT technologies \cite{hadi4}. 

An undeniable part of a smart city is its use of smart agents. These agents can vary a lot in sizes, shapes, and functionalities. They can simply be light sensors that along with their controller act as the energy-saving agents or could be more advanced machines, with complicated controllers and interconnected components that are capable of tackling more advanced problems. The latter agents usually come with an embodiment with numerous sensors and controllers built in them that enable them to perform high-level and human-level tasks such as talking, walking, seeing, and complex reasoning along with the ability to interact with the environment. Embodied Artificial Intelligence is the field of study that takes a deeper look into these agents and explores how they can fit into the real-world and how they can eventually act as our future community workers, personal assistants, robocops, and much more.

Imagine arriving home after a long working day and seeing your home robot waiting for you at the entrance door. Although it is not the most romantic thing ever, you then walk up to it and ask it to make a cup of coffee for you and also add two teaspoons of sugar if there is any in the cabinet. 
For this to become reality, the robot has to have a vast range of skills. It should be able to understand your language and be able to translate questions and instructions to the action. It should be able to see its surroundings and have the ability to recognize objects and scenes. Last but not the least, it must know how to navigate in a big dynamic environment, interact with the objects within it, and be capable of doing long-term planning and reasoning.

In the past few years, there has been significant progress in the fields of computer vision, natural language processing, and reinforcement learning thanks to the advancements in deep learning models. Many things are now possible because of these that seemed impossible a few years ago. However, most of the work has been done in isolation from other lines of work. Meaning that the trained model can only take one type of data (eg. image, text, video) as the input and perform a single task that it is asked for. Consequently, such a model act as a single-sensory machine as opposed to a multi-sensory one. 
Also, for the most part, they all belong to Internet AI rather than Embodied AI. The goal of Internet AI is learning patterns in text, images, and videos from the datasets collected from the internet. 

If we zoom out and look at the way models in Internet AI are being trained, we realize that generally supervised classification is the way to go. For instance, we provide a certain number of dog and cat photos along with the corresponding labels to a perception model and if the number is large enough, the model then can successfully learn the differences that exist between these two animals and discriminate between them. Learning via flashcards falls under the same umbrella for humans.

\begin{figure*}[t]
    \centering
    \includegraphics[width=16.8cm]{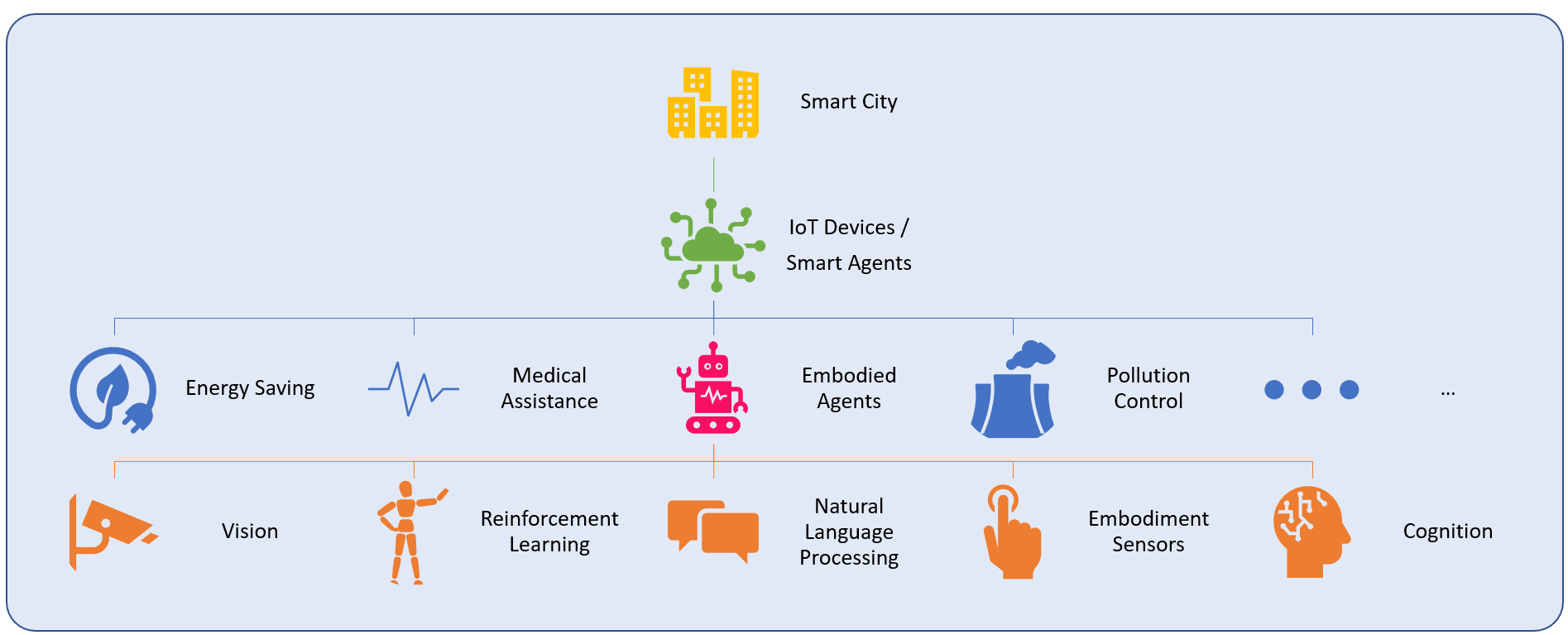}
    \caption{Embodied AI in Smart Cities}
    \label{fig:Fig1}
\end{figure*}

Extensive amount of time has been devoted in the past years to gather and build huge datasets for the imaging and language communities. A few considerable markers of this can be IMAGENET \cite{deng2009imagenet}, MS COCO \cite{lin2014microsoft}, Sun \cite{xiao2010sun}, Caltech-256 \cite{griffin2007caltech},  Places \cite{zhou2014learning} created for vision tasks; Squad \cite{rajpurkar2016squad}, Glue \cite{wang2018glue}, Swag \cite{zellers2018swag} built for language objectives; and also Visual Genome \cite{krishna2017visual} and VQA \cite{antol2015vqa} datasets created for joint purposes to name a few.

Apart from playing a pivotal role in the recent advances of the main fields, these datasets also proved to be useful when used with transfer learning methods to help underlying disciplines such as biomedical imaging \cite{shenavarmasouleh2020drdr, shenavarmasouleh2020drdr2, shenavarmasouleh2021drdrv3}. However, the aforementioned datasets are prune to restrictions. Firstly, at times it can get extremely costly, both in terms of time and money, to gather all the required data for the collection and label them. Secondly, the collection has to be monitored constantly to assure that they follow certain rules to avoid creating biases that could lead to erroneous results in future works \cite{farzan2019} and also make sure that the collected data are all normal and uniform in terms of attributes such as background, size, position of the objects, lighting conditions, etc. while in contrast, we know that in real-world scenarios this cannot be the case and robots have to deal with a mixture of unnormalized noisy irrelevant data plus the relevant well-curated ones. Additionally, the agent would be able to interact with the objects in the wild (e.g. picking it up and looking at the object from another angle) and also use its other senses such as smell and hearing to collect information.

Us humans, we do learn from interactions and it's a must for true intelligence in the real world. In fact, it's not only humans and all the other animals do the same. In, kitten carousel experiment \cite{held1963movement}, Held and Hein exhibited this beautifully. They studied the visual development of two kittens in a carousel over time. One of which had the ability to touch the ground and control its motions within the restrictions of the device while the other was just a passive observer. At the end of the experiment, they found out that the visual development of the former kitten was normal whereas for the latter one it was not, even though they both saw the same thing. This proves that being able to physically experience the world and interact with it is a key element for learning \cite{moravec1984locomotion}.

The goal of Embodied AI is to bring the ability to interact and being able to use multi senses simultaneously into play to enable the robot to continuously learn in a lightly supervised or even unsupervised way in a rich dynamic environment.

\section{Rise of the Embodied AI}
In the mid-1980s a major paradigm shift took place towards embodiment and computer science started to become more practical than theoretical algorithms and approaches. Embedded systems started to appear in all kinds of forms to aid humans in everyday life. Controllers for trains, airplanes, elevators, air conditioners, and Softwares for translation and audio manipulation are some of the most important ones to name a few \cite{hoffmann2012implications}.

Embodied Artificial Intelligence is a broad term, and those successes were for sure great ones to start with; yet, it could clearly be seen that it was a huge room for improvement. Theoretically, the ultimate goal of AI is not only to master any given algorithm or task that is given to, but also gain the ability to multitask and get to human-level intelligence, and that as mentioned requires meaningful interaction with the real world. There are many specialized robots for a vast set of tasks out there, especially in large industries, which can do the assigned task to them to perfection, let it be cutting different metals, painting, soldering circuits, and much more, but until one single machine emerges to have the ability to do different tasks or at least a small subset of them by itself and not just by following orders, it cannot be called intelligence.

Humanoids are the main thing that comes to mind when we talk about robots with intelligence. Although it is the ultimate goal, it is not the only form of intelligence on the earth. Other animals, such as insects have their own kind of intelligence and due to being relatively simpler in comparison to humans, they are a very good place, to begin with.

Rodney Brooks has a famous argument that says it took the evolution much longer to create insects from scratch than getting to human-level intelligence from there. Consequently, he suggested that these simpler biorobotics should be first dealt with in the road to make much more complex ones. Genghis, a six-legged walking robot \cite{brooks1991new} is one of his contributions to this field.

This line of thought was a fundamental change and led researchers to have a change of direction in their work and with that came attention to new domains and topics such as robotics, locomotion, artificial life, bio-inspired systems, and much more. The classical approach did not care about tasks related to interaction with the real world and consequently, locomotion and grasping were the ones to start the journey with.

Since not much computational power was available at the time of this shift, a big challenge for the researchers was the trade-off between simplicity and the potential to operate in complex environments. An extensive amount of work has been done in this area to explore or invent ways to exploit natural body dynamics, materials used in the modules, and their morphologies to make the robots move and become able to grasp and manipulate items without sophisticated processing units \cite{collins2001three, iida2004cheap, yamamoto2001harnessing}. It goes without saying that the ones who could use the physical properties of themselves and the environment to function were more energy-efficient, but they had their own limitations. Not being able to generalize well to complex environments was a major drawback. But, they were fast as the machines with huge processing units needed a reasonable amount of time to think and plan their next action and often move their rigid and non-smooth actuators.

Nowadays, a big part of these issues are solved and we can see extremely fast and smooth natural moving robots capable of doing different types of maneuvers \cite{bledt2018cheetah}, but yet it is foreseen that with the advances of artificial muscles, joints, and tendons this progress can be further improved.

\section{\textbf{Breakdown of Embodied AI}}
In this section, we try to categorize a broad range of research that has been done under the field of Embodied AI. Due to the huge diversity, each section will necessarily be abstract, selective, and reflect the authors’ personal opinion.

\subsection{Language Grounding} 
Machine and human communication has always been a topic of interest. As time goes on, more and more aspects of our lives are controlled by AIs, and hence it is crucial to have ways to talk with them. This is a must for giving new instructions to them or receiving an answer from them, and since we are talking about general day to day machines, we desire this interface to be higher level than programming languages and closer to spoken language. To achieve this, machines must be capable of relating language to actions and the world. 
Language grounding is the field that tries to tackle this and map natural language instructions to robot behavior. 

Hermann et al.'s work show that this can be achieved by rewarding an agent upon successful execution of written instructions in a 3D environment with a combination of unsupervised learning and reinforcement learning \cite{hermann2017grounded}. They also argue that their agent can generalize well after training and can interpret new unseen instructions and operate in unfamiliar situations.

\subsection{Language plus Vision}
Now that we know that machines can understand languages and there exist sophisticated models just for this purpose out there \cite{tenney2019bert}, it is time to bring another sense into play. One of the most popular ways to show the potential of joint training of vision and language is the image and video captioning \cite{pan2017video, ss9281287, title2020, gao2017video, yang2018video}.

More recently, a new line of work has been introduced to take advantage of this connection. Visual Question Answering (VQA) \cite{antol2015vqa} is the task of receiving an image along with a natural language question about that image as an input and attempting to find the accurate natural language answer for it as the output. The beauty of this task is that both the questions and the answers can be open-ended and also the questions can target different aspects of the image such as the objects that are present in them, their relationship or relative positions, colors, and background. 

Following this research, Singh et al. \cite{singh2019towards} cleverly added an OCR module to the VQA model to enable the agent to read the texts available in the image as well and answer questions asked from them or use the additional context indirectly to answer the question better.

One may ask where does the new task stands relative to the previous one. Do agents who can answer questions more intelligent than the ones who deal with captions or not? The answer is yes. In \cite{antol2015vqa} the authors show that VQA agents need a deeper and more detailed understanding of the image and reasoning than models for captioning.

\subsection{Embodied Visual Recognition} 
Passive or fixed agents may fail to recognize objects in scenes if they are partially or heavily occluded. Embodiment comes to the rescue here and gifts the possibility of moving in the environment to actively control the viewing position and angle to remove any ambiguity in object shapes and semantics.

Jayaraman et al. \cite{jayaraman2017learning} started to learn representations that will exploit the link between how the agent moves and how it will affect its visual surrounding. To do this they used raw unlabeled videos along with an external GPS sensor that provided the agent's coordinates and trained their model to learn a representation linking these two. So, after this, the agent would have the ability to predict the outcome of its future actions and guess how the scene would look like after moving forward or turning to a side. 

This was powerful and in a sense, the agent developed imagination. But, there was an issue here. If we pay attention we realize that the agent is still being fed pre-recorded video as the input and is learning similar to the observer kitten in the kitten carousel experiment explained above. So, following this, the authors went after this problem and proposed to train an agent that takes any given object from an arbitrary angle and then predict or better to say imagine the other views by finding the representation in a self-supervised manner \cite{jayaraman2018shapecodes}. 

Up until this point, the agent does not use the sound of its surroundings while humans are all about experiencing the world in a multi-sensory manner. We can see, hear, smell, touch all at the same time, and extract and use the relevant information that could be beneficial to our task at hand. 
All that said, understanding and learning the sound of objects present in a scene is not easy since all the sounds are overlapped and are being received via a single channel sensor. This is often dealt with as an audio source separation problem and lots of work has been done on it in the literature \cite{gao2018learning, parekh2017guiding, pu2017audio, parekh2017motion, Asali2020DeepMSRFAN}.

Now it was the reinforcement learning turn to make a difference. Policies have to be learned to aid agents move around a scene and this is the task of active recognition \cite{aloimonos1988active, ballard1991animate, ballard1992principles, bajcsy1988active, roy2004active}).
The policy will be learned at the same time it is learning other tasks and representation and it will tell the agent where and how to strategically move to recognize things faster \cite{tung2019learning, jayaraman2018end}. 

Results show that policies indeed help the agent to achieve better visual recognition performance and the agents can strategize their future moves and path for better results that are mostly different from shortest paths \cite{yang2019embodied}.


\subsection{Embodied Question Answering} 
Embodied Question Answering brings QA into the embodied world. The task starts by an agent being spawned at a random location in a 3D environment and asked a question which its answer can be found somewhere in the environment. In order for the agent to answer it, it must first strategically navigate to explore the environment, gathers necessary data via its vision, and then answer the question when the agent finds it \cite{das2018embodied, wijmans2019embodied}.

Following this, Das et al. \cite{das2018neural} also presented a modular approach to further enhance this process by teaching the agent to break the master policy into sub-goals that are also interpretable by humans and execute them to answer the question. This proved to increase the success rate.

\subsection{Interactive Question Answering}
Interactive Question Answering (IQA) is closely related to the Embodied version of it. The only main issue is that question is designed in a way that the agent must interact with the environment to find the answer. For example, it has to open the refrigerator, or pick up something from the cabinet and then and plan for a series of actions conditioned on the question \cite{gordon2018iqa}.

\subsection{Multi-Agent Systems}
Multi-Agent Systems (MAS) is another interesting line of development.
The default standpoint of AI has a strong focus on individual agents.
MAS research which has its origins in the field of biology tries to change this and studies the emergence of behaviors in groups of agents or swarms instead \cite{hinchey2007swarms, wang2018bandwidth}.

Every agent has a set of abilities and is good in them to an extent. The point of interest in MAS is how a sophisticated global behavior can emerge from a population of agents working together. A real-life example of such behavior can be found in insects like ants and bees \cite{camazine1999house, langton1995artificial}. One of the interesting goals of this research is to ultimately make agents that could self-repair \cite{hara2003morpho, murata2004self}.

The emerging behavior of MAS can be tailored by researchers to let the group of agents tackle various tasks such as rescue missions, traffic control, fun sports events, surveillance, and much more. Additionally, when fused with other fields unexpected outcomes can occur. Take for instance “Talking Heads” experiment by Luc Steels \cite{steels2001language, steels2003evolving} that showed a common vocabulary emerges through the interaction of agents with each other and their environment via a language game.

\section{\textbf{Simulators}}
Now that we know about the fields and tasks that Embodied AI can shine in, the question is how our agents should be trained. One may say it's good to directly train in the physical world and expose them to its richness. Although a valid solution, this choice comes with a few drawbacks. First, The training process in the real-world is slow, and the process cannot be sped up or parallelized. Second, it is very hard to control the environment and create custom scenarios. Third, it's expensive, both in terms of power and time. Fourth, it's not safe, and improperly trained or not fully trained robots can hurt themselves, humans, animals, and other assets. Fifth, in order for the agent to generalize the training, has to be done in plenty of different environments that is not feasible in this case.

Our next choice is simulators, which can successfully deal with all the aforementioned problems pretty well. In the shift from Internet AI to Embodied AI, simulators take the role that was previously played by traditional datasets. Additionally, one more advantage of using simulators is that the physics in the environment can be tweaked as well. For instance, some traditional approaches in this field \cite{durrant2006simultaneous} are sensitive to noise and for the remedy, the noise in the sensors can be turned off for the purpose of this task.

As a result, agents nowadays are often developed and benchmarked in simulators \cite{gupta2017cognitive, zhu2017target} and once a promising model has been trained and tested, it can then be transferred to the physical world \cite{pomerleau1989alvinn, sadeghi2016cad2rl}.

House3D \cite{wu2018building}, AI2-THOR \cite{kolve2017ai2}, Gibson \cite{xia2018gibson}, CHALET \cite{yan2018chalet}, MINOS \cite{savva2017minos} and Habitat \cite{savva2019Habitat} are some of the popular simulators for the Embodied AI studies. These platforms vary with respect to the 3D environments they use, the tasks they can handle, and the evaluation protocols they provide. These simulators support different sensors such as vision, depth, touch, and semantic segmentation.

In this paper we mainly focus on MINOS and Habitat since they provide more customization abilities (number of sensors, their positions, and their parameters) and are implemented in a loosely coupled manner to generalize well to new multi-sensory tasks and environments. As their API can be used to define any high-level task and the material, object clutter variation, and much more can be programmatically configured for the environment. They both support navigation with both continuous and discrete state spaces. 
Also, for the purpose of their benchmarks, all the actuators are noiseless, but they both have the ability to enable noises if desired \cite{datta2020integrating}.

In the last section, we saw numerous task definitions and how they each can be tackled by the agents. So, before jumping into MINOS and Habitat simulators and reviewing them, let's first get more familiarized with the three main goal-directed navigation tasks, namely, PointGoal Navigation, ObjectGoal Navigation, and RoomGoal Navigation.

In PointGoal Navigation, an agent is appeared at a random starting position and orientation in a 3D environment and is asked to navigate to target coordinates which are given relative to the agent’s position. The agent can access its position via an indoor GPS. There exists no ground-truth map and the agent must only use its sensors to do the task.
The scenarios start the same for ObjectGoal Navigation, and RoomGoal Navigation as well, however, instead of coordinates, the agent is asked to find an object or go to a specific room.

\subsection{MINOS}
Minos simulator provides access to 45,000 three-dimensional models of furnished houses with more than 750K rooms of different types available in the SUNCG \cite{song2017semantic} dataset and 90 multi-floor residences with approximately 2,000 annotated room regions that are in the Matterport3D \cite{chang2017matterport3d} dataset by default. Environments in Matterport3D are more realistic looking than the ones in SUNCG. MINOS simulator can approximately reach hundreds of frames per second on a normal workstation. 

In order to benchmark the system, the authors studied four navigation algorithms; three of which were based on asynchronous advantage actor-critic (A3C) approach \cite{jaderberg2016reinforcement} and the remaining one was Direct Future Prediction (DFP) \cite{dosovitskiy2016learning}.

The most basic one among the algorithms was Feedforward A3C. In this algorithm, a feedforward CNN model is employed as the function approximator to learn the policy along with the total value function that is the expected sum of rewards from the current timestamp until the end of the episode. The second one was LSTM A3C that used an LSTM model with the Feedforward A3C act as a simple memory.
Next was UNREAL, an LSTM A3C model boosted with auxiliary tasks such as value function replay and reward prediction.
Last but not the least, the DFP algorithm was employed that can be considered as Monte Carlo RL \cite{singh1996reinforcement} with a decomposed reward.

The authors benchmarked these algorithms on PointGoal and RoomGoal tasks and found out that firstly, the naive feedforward algorithm fails to learn any useful representation; secondly, in small environments, DFP performs better while in big and more complex environments UNREAL beat the others.

\subsection{Habitat}
Habitat was designed and built in a way to provide the maximum customizability in terms of the datasets that can be used and how the agents and the environment can be configured. That being said, Habitat works with all the major 3D environment datasets without a problem. Moreover, it's extremely fast in comparison to other simulators. 
AI2-THOR and CHALET can get to an fps of roughly ten, 
MINOS and Gibson can get to around a hundred, and House3D yields 300 fps in the best case, while Habitat is capable of getting up to 10,000 frames per second. Habitat also provides a more realistic collision model in which if a collision happens, the agent can be moved partially or not at all in the intended direction.

To benchmark Habitat, the owners employed a few naive algorithm baselines, Proximal Policy Optimization (PPO) \cite{schulman2017proximal} as the representer of learning algorithms versus ORB-SLAM2 \cite{Mur_Artal_2017, mishkin2019benchmarking} as the chosen candidate for non-learning agents and tested them on the PointGoal Navigation task on Gibson and Matterport3D. They used Success weighted by Path Length (SPL) \cite{anderson2018evaluation} as the metric for their performance. The PPO agent was tested with different levels of sensors (e.g. No visual sensor, only depth, only RGB, and RGBD) to perform an ablation study and find the proportion in which each sensor helps the progress. SLAM agents were given RGBD sensors in all the episodes.

The authors found out that first, PPO agents with only RGB perform as bad as agents with no visual sensors. Second, all agents perform better and generalize more on Gibson rather than Matterport3D since the size of environments in the latter is bigger. Third, agents with only depth sensors generalize across datasets the best and can achieve the highest SPL. But most importantly, they realized that unlike what has been mentioned in the previous work, if the PPO agent learns long enough, it will eventually outperform the traditional SLAM pipeline. This finding was only possible because the Habitat simulator was fast enough to train PPO agents for 75million time steps as opposed to only 5million time steps in the previous investigations.


\section{\textbf{Future of Embodied AI}}

\subsection{\textbf{Higher Intelligence}}
Consciousness has always been considered as the ultimate characteristic for true intelligence. Qualia \cite{jackson1982epiphenomenal, tye1997qualia} 
is the philosophical view of consciousness and it is related to the subjective sensory qualities like "the redness of red" that humans have in their mind. If at some point machines can understand this concept and objectively measure such things, then the ultimate goal can be marked as accomplished.

Robots still struggle at performing a wide spectrum of tasks effortlessly and smoothly, and this mainly due to actuator technology as currently mostly electrical motors are used. Advances in artificial muscles and skin sensors that could cover the entire embodiment of the agent would be essential to fully mitigate the human experience in the real world and eventually unlock the desired cognition \cite{hosoda2004robot}.

\subsection{\textbf{Evolution}}
One more key component for cognition is the ability to grow and evolve over time \cite{floreano2004evolution, floreano2008evolutionary, pfeifer2004embodied}. It's easy to evolve the agent's controller via an evolutionary algorithm but it's not enough. If we aim to have completely different agents, we might as well give them the ability to evolve in terms of embodiment and the sensors as well. This again requires the above mentioned artificial cell organism to encode different physical attributes in them and flip them slightly over time. Of course, we are far from this to become reality, but it is always good to know the furthermost step that has to be done one day.

\section{\textbf{Conclusion}}
Embodied AI is the field of study that takes us one step closer to the true intelligence. It is a shift from Internet AI towards embodiment intelligence that tries to exploit the multi-sensory abilities of agents such as vision, hearing, touch, and together with language understanding and reinforcement learning attempts to interact with real-world in a more sensible way.
In this paper, we tried to do a concise review of this field, and its current advancements, subfields, and tools hoping that this would help and accelerate future researches in this area.

\bibliographystyle{unsrt}  
\bibliography{references}  

\begin{thebibliography}{10}

\bibitem{park2021emerging}
Jong~Hyuk Park, Muhammad Younas, Hamid~R Arabnia, and Naveen Chilamkurti.
\newblock Emerging ict applications and services—big data, iot, and cloud
  computing, 2021.

\bibitem{hadi2_interdependent}
M~Hadi Amini, Ahmed Imteaj, and Panos~M Pardalos.
\newblock Interdependent networks: A data science perspective.
\newblock {\em Patterns}, page 100003, 2020.

\bibitem{mohammadi2019promises}
Farid~Ghareh Mohammadi and M~Hadi Amini.
\newblock Promises of meta-learning for device-free human sensing: learn to
  sense.
\newblock In {\em Proceedings of the 1st ACM International Workshop on
  Device-Free Human Sensing}, pages 44--47, 2019.

\bibitem{hadi3_distributed}
M~Hadi Amini, Javad Mohammadi, and Soummya Kar.
\newblock Promises of fully distributed optimization for iot-based smart city
  infrastructures.
\newblock In {\em Optimization, Learning, and Control for Interdependent
  Complex Networks}, pages 15--35. Springer, 2020.

\bibitem{hadi5_sustainable}
M~Hadi Amini, Hamidreza Arasteh, and Pierluigi Siano.
\newblock Sustainable smart cities through the lens of complex interdependent
  infrastructures: panorama and state-of-the-art.
\newblock In {\em Sustainable interdependent networks II}, pages 45--68.
  Springer, 2019.

\bibitem{hadi1_pollution}
Ditsuhi Iskandaryan, Francisco Ramos, and Sergio Trilles.
\newblock Air quality prediction in smart cities using machine learning
  technologies based on sensor data: A review.
\newblock {\em Applied Sciences}, 10(7):2401, 2020.

\bibitem{hadi4}
Michael Batty, Kay~W Axhausen, Fosca Giannotti, Alexei Pozdnoukhov, Armando
  Bazzani, Monica Wachowicz, Georgios Ouzounis, and Yuval Portugali.
\newblock Smart cities of the future.
\newblock {\em The European Physical Journal Special Topics}, 214(1):481--518,
  2012.

\bibitem{deng2009imagenet}
Jia Deng, Wei Dong, Richard Socher, Li-Jia Li, Kai Li, and Li~Fei-Fei.
\newblock Imagenet: A large-scale hierarchical image database.
\newblock In {\em 2009 IEEE conference on computer vision and pattern
  recognition}, pages 248--255. Ieee, 2009.

\bibitem{lin2014microsoft}
Tsung-Yi Lin, Michael Maire, Serge Belongie, James Hays, Pietro Perona, Deva
  Ramanan, Piotr Doll{\'a}r, and C~Lawrence Zitnick.
\newblock Microsoft coco: Common objects in context.
\newblock In {\em European conference on computer vision}, pages 740--755.
  Springer, 2014.

\bibitem{xiao2010sun}
Jianxiong Xiao, James Hays, Krista~A Ehinger, Aude Oliva, and Antonio Torralba.
\newblock Sun database: Large-scale scene recognition from abbey to zoo.
\newblock In {\em 2010 IEEE computer society conference on computer vision and
  pattern recognition}, pages 3485--3492. IEEE, 2010.

\bibitem{griffin2007caltech}
Gregory Griffin, Alex Holub, and Pietro Perona.
\newblock Caltech-256 object category dataset.
\newblock 2007.

\bibitem{zhou2014learning}
Bolei Zhou, Agata Lapedriza, Jianxiong Xiao, Antonio Torralba, and Aude Oliva.
\newblock Learning deep features for scene recognition using places database.
\newblock {\em Advances in neural information processing systems}, 27:487--495,
  2014.

\bibitem{rajpurkar2016squad}
Pranav Rajpurkar, Jian Zhang, Konstantin Lopyrev, and Percy Liang.
\newblock Squad: 100,000+ questions for machine comprehension of text.
\newblock {\em arXiv preprint arXiv:1606.05250}, 2016.

\bibitem{wang2018glue}
Alex Wang, Amanpreet Singh, Julian Michael, Felix Hill, Omer Levy, and Samuel~R
  Bowman.
\newblock Glue: A multi-task benchmark and analysis platform for natural
  language understanding.
\newblock {\em arXiv preprint arXiv:1804.07461}, 2018.

\bibitem{zellers2018swag}
Rowan Zellers, Yonatan Bisk, Roy Schwartz, and Yejin Choi.
\newblock Swag: A large-scale adversarial dataset for grounded commonsense
  inference.
\newblock {\em arXiv preprint arXiv:1808.05326}, 2018.

\bibitem{krishna2017visual}
Ranjay Krishna, Yuke Zhu, Oliver Groth, Justin Johnson, Kenji Hata, Joshua
  Kravitz, Stephanie Chen, Yannis Kalantidis, Li-Jia Li, David~A Shamma, et~al.
\newblock Visual genome: Connecting language and vision using crowdsourced
  dense image annotations.
\newblock {\em International journal of computer vision}, 123(1):32--73, 2017.

\bibitem{antol2015vqa}
Stanislaw Antol, Aishwarya Agrawal, Jiasen Lu, Margaret Mitchell, Dhruv Batra,
  C~Lawrence~Zitnick, and Devi Parikh.
\newblock Vqa: Visual question answering.
\newblock In {\em Proceedings of the IEEE international conference on computer
  vision}, pages 2425--2433, 2015.

\bibitem{shenavarmasouleh2020drdr}
Farzan Shenavarmasouleh and Hamid~R Arabnia.
\newblock Drdr: Automatic masking of exudates and microaneurysms caused by
  diabetic retinopathy using mask r-cnn and transfer learning.
\newblock {\em arXiv preprint arXiv:2007.02026}, 2020.

\bibitem{shenavarmasouleh2020drdr2}
Farzan Shenavarmasouleh, Farid~Ghareh Mohammadi, M~Hadi Amini, and Hamid~R
  Arabnia.
\newblock Drdr ii: Detecting the severity level of diabetic retinopathy using
  mask rcnn and transfer learning.
\newblock {\em arXiv preprint arXiv:2011.14733}, 2020.

\bibitem{shenavarmasouleh2021drdrv3}
Farzan Shenavarmasouleh, Farid~Ghareh Mohammadi, M.~Hadi Amini, Thiab Taha,
  Khaled Rasheed, and Hamid~R. Arabnia.
\newblock Drdrv3: Complete lesion detection in fundus images using mask r-cnn,
  transfer learning, and lstm.
\newblock {\em arXiv preprint arXiv:2108.08095}, 2021.

\bibitem{farzan2019}
F.~{Shenavarmasouleh} and H.~{Arabnia}.
\newblock Causes of misleading statistics and research results
  irreproducibility: A concise review.
\newblock In {\em 2019 International Conference on Computational Science and
  Computational Intelligence (CSCI)}, pages 465--470, 2019.

\bibitem{held1963movement}
Richard Held and Alan Hein.
\newblock Movement-produced stimulation in the development of visually guided
  behavior.
\newblock {\em Journal of comparative and physiological psychology}, 56(5):872,
  1963.

\bibitem{moravec1984locomotion}
Hans Moravec.
\newblock Locomotion, vision and intelligence.
\newblock 1984.

\bibitem{hoffmann2012implications}
Matej Hoffmann and Rolf Pfeifer.
\newblock The implications of embodiment for behavior and cognition: animal and
  robotic case studies.
\newblock {\em arXiv preprint arXiv:1202.0440}, 2012.

\bibitem{brooks1991new}
Rodney~A Brooks.
\newblock New approaches to robotics.
\newblock {\em Science}, 253(5025):1227--1232, 1991.

\bibitem{collins2001three}
Steven~H Collins, Martijn Wisse, and Andy Ruina.
\newblock A three-dimensional passive-dynamic walking robot with two legs and
  knees.
\newblock {\em The International Journal of Robotics Research}, 20(7):607--615,
  2001.

\bibitem{iida2004cheap}
Fumiya Iida and Rolf Pfeifer.
\newblock Cheap rapid locomotion of a quadruped robot: Self-stabilization of
  bounding gait.
\newblock In {\em Intelligent autonomous systems}, volume~8, pages 642--649.
  IOS Press Amsterdam, The Netherlands, 2004.

\bibitem{yamamoto2001harnessing}
Tomoyuki Yamamoto and Yasuo Kuniyoshi.
\newblock Harnessing the robot's body dynamics: a global dynamics approach.
\newblock In {\em Proceedings 2001 IEEE/RSJ International Conference on
  Intelligent Robots and Systems. Expanding the Societal Role of Robotics in
  the the Next Millennium (Cat. No. 01CH37180)}, volume~1, pages 518--525.
  IEEE, 2001.

\bibitem{bledt2018cheetah}
Gerardo Bledt, Matthew~J Powell, Benjamin Katz, Jared Di~Carlo, Patrick~M
  Wensing, and Sangbae Kim.
\newblock Mit cheetah 3: Design and control of a robust, dynamic quadruped
  robot.
\newblock In {\em 2018 IEEE/RSJ International Conference on Intelligent Robots
  and Systems (IROS)}, pages 2245--2252. IEEE, 2018.

\bibitem{hermann2017grounded}
Karl~Moritz Hermann, Felix Hill, Simon Green, Fumin Wang, Ryan Faulkner, Hubert
  Soyer, David Szepesvari, Wojciech~Marian Czarnecki, Max Jaderberg, Denis
  Teplyashin, et~al.
\newblock Grounded language learning in a simulated 3d world.
\newblock {\em arXiv preprint arXiv:1706.06551}, 2017.

\bibitem{tenney2019bert}
Ian Tenney, Dipanjan Das, and Ellie Pavlick.
\newblock Bert rediscovers the classical nlp pipeline, 2019.

\bibitem{pan2017video}
Yingwei Pan, Ting Yao, Houqiang Li, and Tao Mei.
\newblock Video captioning with transferred semantic attributes.
\newblock In {\em Proceedings of the IEEE conference on computer vision and
  pattern recognition}, pages 6504--6512, 2017.

\bibitem{ss9281287}
Soheyla Amirian, Khaled Rasheed, Thiab~R. Taha, and Hamid~R. Arabnia.
\newblock Automatic image and video caption generation with deep learning: A
  concise review and algorithmic overlap.
\newblock {\em IEEE Access}, 8:218386--218400, 2020.

\bibitem{title2020}
Soheyla Amirian, Khaled Rasheed, Thiab~R. Taha, and Hamid~R. Arabnia.
\newblock Automatic generation of descriptive titles for video clips using deep
  learning.
\newblock In {\em Springer Nature - Research Book Series:Transactions on
  Computational Science \& Computational Intelligence}, page Springer ID:
  89066307, 2020.

\bibitem{gao2017video}
Lianli Gao, Zhao Guo, Hanwang Zhang, Xing Xu, and Heng~Tao Shen.
\newblock Video captioning with attention-based lstm and semantic consistency.
\newblock {\em IEEE Transactions on Multimedia}, 19(9):2045--2055, 2017.

\bibitem{yang2018video}
Yang Yang, Jie Zhou, Jiangbo Ai, Yi~Bin, Alan Hanjalic, Heng~Tao Shen, and
  Yanli Ji.
\newblock Video captioning by adversarial lstm.
\newblock {\em IEEE Transactions on Image Processing}, 27(11):5600--5611, 2018.

\bibitem{singh2019towards}
Amanpreet Singh, Vivek Natarajan, Meet Shah, Yu~Jiang, Xinlei Chen, Dhruv
  Batra, Devi Parikh, and Marcus Rohrbach.
\newblock Towards vqa models that can read.
\newblock In {\em Proceedings of the IEEE Conference on Computer Vision and
  Pattern Recognition}, pages 8317--8326, 2019.

\bibitem{jayaraman2017learning}
Dinesh Jayaraman and Kristen Grauman.
\newblock Learning image representations tied to egomotion from unlabeled
  video.
\newblock {\em International Journal of Computer Vision}, 125(1-3):136--161,
  2017.

\bibitem{jayaraman2018shapecodes}
Dinesh Jayaraman, Ruohan Gao, and Kristen Grauman.
\newblock Shapecodes: self-supervised feature learning by lifting views to
  viewgrids.
\newblock In {\em Proceedings of the European Conference on Computer Vision
  (ECCV)}, pages 120--136, 2018.

\bibitem{gao2018learning}
Ruohan Gao, Rogerio Feris, and Kristen Grauman.
\newblock Learning to separate object sounds by watching unlabeled video.
\newblock In {\em Proceedings of the European Conference on Computer Vision
  (ECCV)}, pages 35--53, 2018.

\bibitem{parekh2017guiding}
Sanjeel Parekh, Slim Essid, Alexey Ozerov, Ngoc~QK Duong, Patrick P{\'e}rez,
  and Ga{\"e}l Richard.
\newblock Guiding audio source separation by video object information.
\newblock In {\em 2017 IEEE Workshop on Applications of Signal Processing to
  Audio and Acoustics (WASPAA)}, pages 61--65. IEEE, 2017.

\bibitem{pu2017audio}
Jie Pu, Yannis Panagakis, Stavros Petridis, and Maja Pantic.
\newblock Audio-visual object localization and separation using low-rank and
  sparsity.
\newblock In {\em 2017 IEEE International Conference on Acoustics, Speech and
  Signal Processing (ICASSP)}, pages 2901--2905. IEEE, 2017.

\bibitem{parekh2017motion}
Sanjeel Parekh, Slim Essid, Alexey Ozerov, Ngoc~QK Duong, Patrick P{\'e}rez,
  and Ga{\"e}l Richard.
\newblock Motion informed audio source separation.
\newblock In {\em 2017 IEEE International Conference on Acoustics, Speech and
  Signal Processing (ICASSP)}, pages 6--10. IEEE, 2017.

\bibitem{Asali2020DeepMSRFAN}
Ehsan Asali, Farzan Shenavarmasouleh, F.~Mohammadi, P.~Suresh, and H.~Arabnia.
\newblock Deepmsrf: A novel deep multimodal speaker recognition framework with
  feature selection.
\newblock {\em ArXiv}, abs/2007.06809, 2020.

\bibitem{aloimonos1988active}
John Aloimonos, Isaac Weiss, and Amit Bandyopadhyay.
\newblock Active vision.
\newblock {\em International journal of computer vision}, 1(4):333--356, 1988.

\bibitem{ballard1991animate}
Dana~H Ballard.
\newblock Animate vision.
\newblock {\em Artificial intelligence}, 48(1):57--86, 1991.

\bibitem{ballard1992principles}
Dana~H Ballard and Christopher~M Brown.
\newblock Principles of animate vision.
\newblock {\em CVGIP: Image Understanding}, 56(1):3--21, 1992.

\bibitem{bajcsy1988active}
Ruzena Bajcsy.
\newblock Active perception.
\newblock {\em Proceedings of the IEEE}, 76(8):966--1005, 1988.

\bibitem{roy2004active}
Sumantra~Dutta Roy, Santanu Chaudhury, and Subhashis Banerjee.
\newblock Active recognition through next view planning: a survey.
\newblock {\em Pattern Recognition}, 37(3):429--446, 2004.

\bibitem{tung2019learning}
Hsiao-Yu~Fish Tung, Ricson Cheng, and Katerina Fragkiadaki.
\newblock Learning spatial common sense with geometry-aware recurrent networks.
\newblock In {\em Proceedings of the IEEE Conference on Computer Vision and
  Pattern Recognition}, pages 2595--2603, 2019.

\bibitem{jayaraman2018end}
Dinesh Jayaraman and Kristen Grauman.
\newblock End-to-end policy learning for active visual categorization.
\newblock {\em IEEE transactions on pattern analysis and machine intelligence},
  41(7):1601--1614, 2018.

\bibitem{yang2019embodied}
Jianwei Yang, Zhile Ren, Mingze Xu, Xinlei Chen, David Crandall, Devi Parikh,
  and Dhruv Batra.
\newblock Embodied visual recognition, 2019.

\bibitem{das2018embodied}
Abhishek Das, Samyak Datta, Georgia Gkioxari, Stefan Lee, Devi Parikh, and
  Dhruv Batra.
\newblock Embodied question answering.
\newblock In {\em Proceedings of the IEEE Conference on Computer Vision and
  Pattern Recognition Workshops}, pages 2054--2063, 2018.

\bibitem{wijmans2019embodied}
Erik Wijmans, Samyak Datta, Oleksandr Maksymets, Abhishek Das, Georgia
  Gkioxari, Stefan Lee, Irfan Essa, Devi Parikh, and Dhruv Batra.
\newblock Embodied question answering in photorealistic environments with point
  cloud perception.
\newblock In {\em Proceedings of the IEEE Conference on Computer Vision and
  Pattern Recognition}, pages 6659--6668, 2019.

\bibitem{das2018neural}
Abhishek Das, Georgia Gkioxari, Stefan Lee, Devi Parikh, and Dhruv Batra.
\newblock Neural modular control for embodied question answering.
\newblock {\em arXiv preprint arXiv:1810.11181}, 2018.

\bibitem{gordon2018iqa}
Daniel Gordon, Aniruddha Kembhavi, Mohammad Rastegari, Joseph Redmon, Dieter
  Fox, and Ali Farhadi.
\newblock Iqa: Visual question answering in interactive environments.
\newblock In {\em Proceedings of the IEEE Conference on Computer Vision and
  Pattern Recognition}, pages 4089--4098, 2018.

\bibitem{hinchey2007swarms}
Michael~G Hinchey, Roy Sterritt, and Chris Rouff.
\newblock Swarms and swarm intelligence.
\newblock {\em Computer}, 40(4):111--113, 2007.

\bibitem{wang2018bandwidth}
Junjue Wang, Ziqiang Feng, Zhuo Chen, Shilpa George, Mihir Bala, Padmanabhan
  Pillai, Shao-Wen Yang, and Mahadev Satyanarayanan.
\newblock Bandwidth-efficient live video analytics for drones via edge
  computing.
\newblock In {\em 2018 IEEE/ACM Symposium on Edge Computing (SEC)}, pages
  159--173. IEEE, 2018.

\bibitem{camazine1999house}
Scott Camazine, Peter~K Visscher, Jennifer Finley, and Richard~S Vetter.
\newblock House-hunting by honey bee swarms: collective decisions and
  individual behaviors.
\newblock {\em Insectes Sociaux}, 46(4):348--360, 1999.

\bibitem{langton1995artificial}
Christopher~G Langton.
\newblock Artificial life: An overview cambridge.
\newblock {\em Mass. MIT}, 1995.

\bibitem{hara2003morpho}
Fumio Hara and Rolf Pfeifer.
\newblock {\em Morpho-functional machines: The new species: Designing embodied
  intelligence}.
\newblock Springer Science \& Business Media, 2003.

\bibitem{murata2004self}
Satoshi Murata, Akiya Kamimura, Haruhisa Kurokawa, Eiichi Yoshida, Kohji
  Tomita, and Shigeru Kokaji.
\newblock Self-reconfigurable robots: Platforms for emerging functionality.
\newblock In {\em Embodied Artificial Intelligence}, pages 312--330. Springer,
  2004.

\bibitem{steels2001language}
Luc Steels.
\newblock Language games for autonomous robots.
\newblock {\em IEEE Intelligent systems}, 16(5):16--22, 2001.

\bibitem{steels2003evolving}
Luc Steels.
\newblock Evolving grounded communication for robots.
\newblock {\em Trends in cognitive sciences}, 7(7):308--312, 2003.

\bibitem{durrant2006simultaneous}
Hugh Durrant-Whyte and Tim Bailey.
\newblock Simultaneous localization and mapping: part i.
\newblock {\em IEEE robotics \& automation magazine}, 13(2):99--110, 2006.

\bibitem{gupta2017cognitive}
Saurabh Gupta, James Davidson, Sergey Levine, Rahul Sukthankar, and Jitendra
  Malik.
\newblock Cognitive mapping and planning for visual navigation.
\newblock In {\em Proceedings of the IEEE Conference on Computer Vision and
  Pattern Recognition}, pages 2616--2625, 2017.

\bibitem{zhu2017target}
Yuke Zhu, Roozbeh Mottaghi, Eric Kolve, Joseph~J Lim, Abhinav Gupta,
  Li~Fei-Fei, and Ali Farhadi.
\newblock Target-driven visual navigation in indoor scenes using deep
  reinforcement learning.
\newblock In {\em 2017 IEEE international conference on robotics and automation
  (ICRA)}, pages 3357--3364. IEEE, 2017.

\bibitem{pomerleau1989alvinn}
Dean~A Pomerleau.
\newblock Alvinn: An autonomous land vehicle in a neural network.
\newblock In {\em Advances in neural information processing systems}, pages
  305--313, 1989.

\bibitem{sadeghi2016cad2rl}
Fereshteh Sadeghi and Sergey Levine.
\newblock Cad2rl: Real single-image flight without a single real image.
\newblock {\em arXiv preprint arXiv:1611.04201}, 2016.

\bibitem{wu2018building}
Yi~Wu, Yuxin Wu, Georgia Gkioxari, and Yuandong Tian.
\newblock Building generalizable agents with a realistic and rich 3d
  environment.
\newblock {\em arXiv preprint arXiv:1801.02209}, 2018.

\bibitem{kolve2017ai2}
Eric Kolve, Roozbeh Mottaghi, Winson Han, Eli VanderBilt, Luca Weihs, Alvaro
  Herrasti, Daniel Gordon, Yuke Zhu, Abhinav Gupta, and Ali Farhadi.
\newblock Ai2-thor: An interactive 3d environment for visual ai.
\newblock {\em arXiv preprint arXiv:1712.05474}, 2017.

\bibitem{xia2018gibson}
Fei Xia, Amir~R Zamir, Zhiyang He, Alexander Sax, Jitendra Malik, and Silvio
  Savarese.
\newblock Gibson env: Real-world perception for embodied agents.
\newblock In {\em Proceedings of the IEEE Conference on Computer Vision and
  Pattern Recognition}, pages 9068--9079, 2018.

\bibitem{yan2018chalet}
Claudia Yan, Dipendra Misra, Andrew Bennnett, Aaron Walsman, Yonatan Bisk, and
  Yoav Artzi.
\newblock Chalet: Cornell house agent learning environment.
\newblock {\em arXiv preprint arXiv:1801.07357}, 2018.

\bibitem{savva2017minos}
Manolis Savva, Angel~X Chang, Alexey Dosovitskiy, Thomas Funkhouser, and
  Vladlen Koltun.
\newblock Minos: Multimodal indoor simulator for navigation in complex
  environments.
\newblock {\em arXiv preprint arXiv:1712.03931}, 2017.

\bibitem{savva2019Habitat}
Manolis Savva, Abhishek Kadian, Oleksandr Maksymets, Yili Zhao, Erik Wijmans,
  Bhavana Jain, Julian Straub, Jia Liu, Vladlen Koltun, Jitendra Malik, et~al.
\newblock Habitat: A platform for embodied ai research.
\newblock In {\em Proceedings of the IEEE International Conference on Computer
  Vision}, pages 9339--9347, 2019.

\bibitem{datta2020integrating}
Samyak Datta, Oleksandr Maksymets, Judy Hoffman, Stefan Lee, Dhruv Batra, and
  Devi Parikh.
\newblock Integrating egocentric localization for more realistic point-goal
  navigation agents.
\newblock {\em arXiv preprint arXiv:2009.03231}, 2020.

\bibitem{song2017semantic}
Shuran Song, Fisher Yu, Andy Zeng, Angel~X Chang, Manolis Savva, and Thomas
  Funkhouser.
\newblock Semantic scene completion from a single depth image.
\newblock In {\em Proceedings of the IEEE Conference on Computer Vision and
  Pattern Recognition}, pages 1746--1754, 2017.

\bibitem{chang2017matterport3d}
Angel Chang, Angela Dai, Thomas Funkhouser, Maciej Halber, Matthias Niessner,
  Manolis Savva, Shuran Song, Andy Zeng, and Yinda Zhang.
\newblock Matterport3d: Learning from rgb-d data in indoor environments.
\newblock {\em arXiv preprint arXiv:1709.06158}, 2017.

\bibitem{jaderberg2016reinforcement}
Max Jaderberg, Volodymyr Mnih, Wojciech~Marian Czarnecki, Tom Schaul, Joel~Z
  Leibo, David Silver, and Koray Kavukcuoglu.
\newblock Reinforcement learning with unsupervised auxiliary tasks.
\newblock {\em arXiv preprint arXiv:1611.05397}, 2016.

\bibitem{dosovitskiy2016learning}
Alexey Dosovitskiy and Vladlen Koltun.
\newblock Learning to act by predicting the future.
\newblock {\em arXiv preprint arXiv:1611.01779}, 2016.

\bibitem{singh1996reinforcement}
Satinder~P Singh and Richard~S Sutton.
\newblock Reinforcement learning with replacing eligibility traces.
\newblock {\em Machine learning}, 22(1-3):123--158, 1996.

\bibitem{schulman2017proximal}
John Schulman, Filip Wolski, Prafulla Dhariwal, Alec Radford, and Oleg Klimov.
\newblock Proximal policy optimization algorithms.
\newblock {\em arXiv preprint arXiv:1707.06347}, 2017.

\bibitem{Mur_Artal_2017}
Raul Mur-Artal and Juan~D. Tardos.
\newblock Orb-slam2: An open-source slam system for monocular, stereo, and
  rgb-d cameras.
\newblock {\em IEEE Transactions on Robotics}, 33(5):1255–1262, Oct 2017.

\bibitem{mishkin2019benchmarking}
Dmytro Mishkin, Alexey Dosovitskiy, and Vladlen Koltun.
\newblock Benchmarking classic and learned navigation in complex 3d
  environments.
\newblock {\em arXiv preprint arXiv:1901.10915}, 2019.

\bibitem{anderson2018evaluation}
Peter Anderson, Angel Chang, Devendra~Singh Chaplot, Alexey Dosovitskiy,
  Saurabh Gupta, Vladlen Koltun, Jana Kosecka, Jitendra Malik, Roozbeh
  Mottaghi, Manolis Savva, et~al.
\newblock On evaluation of embodied navigation agents.
\newblock {\em arXiv preprint arXiv:1807.06757}, 2018.

\bibitem{jackson1982epiphenomenal}
Frank Jackson.
\newblock Epiphenomenal qualia.
\newblock {\em The Philosophical Quarterly (1950-)}, 32(127):127--136, 1982.

\bibitem{tye1997qualia}
Michael Tye.
\newblock Qualia.
\newblock 1997.

\bibitem{hosoda2004robot}
Koh Hosoda.
\newblock Robot finger design for developmental tactile interaction.
\newblock In {\em Embodied Artificial Intelligence}, pages 219--230. Springer,
  2004.

\bibitem{floreano2004evolution}
Dario Floreano, Francesco Mondada, Andres Perez-Uribe, and Daniel Roggen.
\newblock Evolution of embodied intelligence.
\newblock In {\em Embodied artificial intelligence}, pages 293--311. Springer,
  2004.

\bibitem{floreano2008evolutionary}
Dario Floreano, Phil Husbands, and Stefano Nolfi.
\newblock Evolutionary robotics.
\newblock Technical report, Springer Verlag, 2008.

\bibitem{pfeifer2004embodied}
Rolf Pfeifer and Fumiya Iida.
\newblock Embodied artificial intelligence: Trends and challenges.
\newblock In {\em Embodied artificial intelligence}, pages 1--26. Springer,
  2004.

\end{thebibliography}

\end{document}